# Weighing and Integrating Evidence for Stochastic Simulation in Bayesian Networks

**Robert Fung and Kuo-Chu Chang**
Advanced Decision Systems
1500 Plymouth Street
Mountain View, California 94043-1230

## Abstract

Stochastic simulation approaches perform probabilistic inference in Bayesian networks by estimating the probability of an event based on the frequency that that event occurs in a set of simulation trials. This paper describes the *evidence weighting* mechanism, for augmenting the logic sampling stochastic simulation algorithm [5]. Evidence weighting modifies the logic sampling algorithm by weighting each simulation trial by the likelihood of a network's evidence given the sampled state node values for that trial. We also describe an enhancement to the basic algorithm which uses the evidential integration technique [2]. A comparison of the basic evidence weighting mechanism with the Markov blanket algorithm [8], the logic sampling algorithm, and the evidence integration algorithm is presented. The comparison is aided by analyzing the performance of the algorithms in a simple example network.

## 1 Introduction

One of the newer approaches for inference in Bayesian Network is stochastic simulation. In this approach, the probability of an event of interest is estimated using the frequency that the event occurs in a set of simulation trials. This kind of approximate methods have been recognized to be a valuable tool since it has been shown that exact probabilistic inference is NP-hard [4] in general. Therefore, for networks which cannot be effectively addressed by exact methods [6, 7, 10], approximate inference schemes such as stochastic simulation are the only alternative for making inference computationally feasible.

This paper describes a simple but promising mechanism, evidence weighting, for augmenting the logic sampling algorithm [5]. Logic sampling has been shown to have one major drawback when dealing with evidence. The mechanism described in this paper has some capability to handle this problem without the introduction of other drawbacks (e.g., dealing with deterministic nodes).

When applying the logic sampling algorithm on networks with no evidence, sampling in each trial starts from the root nodes (i.e., nodes with no predecessors) and works down to the leaf nodes. The prior distribution of each root node is used to guide the choice of a sample value from the node's state space. Once a sample value for a node has been chosen, the sample value is "inserted" into the node's successors. The insertion of a value into a successor node "removes" the root node as a predecessor. Because Bayesian Nets are acyclic, once any root node is sampled, it must leave the graph with at least one new "root" node (i.e., a node with no "uninstantiated" predecessors). This property insures that the process of sampling will continue until all nodes in a network are sampled.

At the end of each trial, the count for each event of interest is updated. If the event occurs in the trial, then it is incremented by a constant (e.g., 1). If the event does not occur in the trial then its count is not incremented. The probability of any event of interest represented in the network can be estimated based on its frequency of occurrence (e.g., the event's count divided by the number of trials).

The major disadvantage of the logic sampling approach is that it does not deal well with evidence. When there is evidence in a network, sampling proceeds as described above with the addition of an acceptance procedure. For each simulation run, the acceptance procedure checks whether the sampled values for the evidence nodes match the observed evidence. If the values do not match, the results of that simulation run are discarded. If the values do match, the trial is "valid" and the sampled values are used to update the count of each event of interest. In cases where there are either large numbers of observations or when the *a priori* probability of the observed evidence is low, the percentage of "valid" simulations is low and therefore the number of simulation runs needed may be quite large.

To overcome the disadvantage of the logic sampling algorithm, a "Markov blanket" approach [8] has been proposed which modifies the algorithm by adding a pre-processing step to each simulation trial. This preprocessing step involves each node performing some local computation. This computation involves looking at its Markov blanket to determine a probability distribution for sampling. While the Markov blanket approach overcomes the major disadvantage of the logic sampling approach, by doing so, it has introduced some negative side effects. Most importantly, the convergence rate of this approach deteriorates when deterministic functions or highly dependent nodes are present in a network [2]. Secondly, since the pre-processing step must take place for every node for every trial, the computation needed per trial for this approach is often greater than for the simple logic-sampling approach.

Another approach which has been proposed to overcome the disadvantage of the logic sampling algorithm is "evidential integration" [2]. Evidential integration is a pre-processing step in which the constraints imposed by a network's evidence are integrated into the network using the arc reversal operation. Evidential integration transforms the network by an iterative application of Bayes' Rule. The transformation can be expressed as follows:

$$p(X|E) = kP(E|X)P(X) \quad (1)$$

where X represents the states of the network, $P(X)$ represents the *a priori* joint distribution of the network, $p(E|X)$ represents the assessed likelihood of the evidence given all the states, $p(X|E)$ represents the joint distribution of the new network in which the evidence has been integrated, and $k$ is a normalization constant.

The algorithm for evidential integration is as follows. For every evidence node, reverse arcs from predecessors which are not evidence nodes until the evidence node has no predecessors which are state nodes. Once a network has had this step performed, the logic sampling approach can be applied straightforwardly to estimate the posterior probability of any event of interest. This works because the evidential integration step creates the posterior distribution for the network. However, the evidential integration step may be expensive if the network is dense and heavily connected with the evidences.

In this paper, we present a mechanism which keeps the advantages of the logic-sampling approach while removes its major shortage (i.e., dealing with evidence). In the evidence weighting algorithm, only the state nodes of a network are simulated in each trial. For each simulation trial, the likelihood of all the evidence given the sampled state values is used to increment the count of each event of interest. The estimated probability distribution is obtained by normalizing after all the simulation trials are complete. The primary drawback of the evidence weighting algorithm is that when evidence likelihoods are extremal, the algorithm reduces to logic sampling and may converge slowly. To handle this problem, we propose an extension of the evidence weighting method by incorporating the evidential integration mechanism. Examples and simulation results are also given to compare various algorithms.

This paper is organized as follows. Section 2 describes the evidence weighting method. Section 3 presents an extension of the evidence weighting method modified with the evidential integration mechanism. Section 4 presents the simulation results with the widely used example [3]. A discussion of the algorithms through comparison with the Markov blanket simulation method [8], the logic sampling algorithm, and the evidential integration algorithm are given in Section 5. Some concluding remarks are given in Section 6.

## 2 The Evidence Weighting Technique

In evidence weighting, the logic sampling algorithm is modified by considering a likelihood weight for each trial and sampling only state nodes. For each simulation trial, the likelihood of each piece of evidence given the sampled state node values is found. The product of these likelihoods is then used instead of a constant to increment the count of each event of interest.

Referring back to equation (1), evidence weighting works by restricting sampling to the *a priori* distribution $P(X)$ and then weighting each sample by the weight $P(E|X)$, where $P(E|X)$ can be easily obtained as:

$$P(E|X) = \prod_{E_i \in E} P(E_i|C(E_i)) \quad (2)$$

where $C(E_i)$ are the direct predecessors of $E_i$.

The justification of this procedure is straightforward. By sampling from the *a priori* distribution $P(X)$ and weighting each sample by $P(E|X)$, the *a posterior* probability of an event of interest $y$, $P(y|E)$, can be estimated as below,

$$p(y|E) \approx k\tfrac{1}{N}\textstyle\sum_{i=1}^{N} P(E|x_i)U(x_i) \quad (3)$$

where $x_i$ is the realization of $X$ in the $i-th$ trial, $U(x_i)$ is 1 if $x_i \subset y$ and 0 otherwise, and $N$ is the total number of trials.

The algorithm has what appears to be a substantial advantage over the logic sampling approach since all trials are "valid" and contribute to the reduction of error. However if the likelihoods of all evidence nodes in a network are extreme (i.e., 1 or 0) this advantage will disappear and the evidence weighting mechanism will reduce to logic sampling.

Not all pieces of evidence will bear on all events of interest. The determination of which pieces of evidence bear which events can result in the "sufficient information" [10] needed to perform inference. This algorithm may converge more quickly (in trials) if the evidence weight for each event in the simulation trial only contains the likelihoods from pieces of evidence which bear on that event. This procedure may remove the "noise" created by the nonbearing evidence likelihoods. However we have not studied the tradeoff between the extra computations required and the improvement on convergence rate, if any, to know in what situations this calculation will be useful.

## 3 Evidence Weighting With Evidential Integration

As mentioned above, one major disvantage of the evidence weighting technique is the slow convergence in the situations where the likelihoods of evidence are extremal. One way of handling this is to incorporate the evidential integration mechanism. As described in the introduction, the evidential integration is a pre-processing step in which the evidences are integrated into the network using the arc reversal operation. By doing so, it may be possible to transform the network into one that will generate stochastic sample instances more efficiently [2].

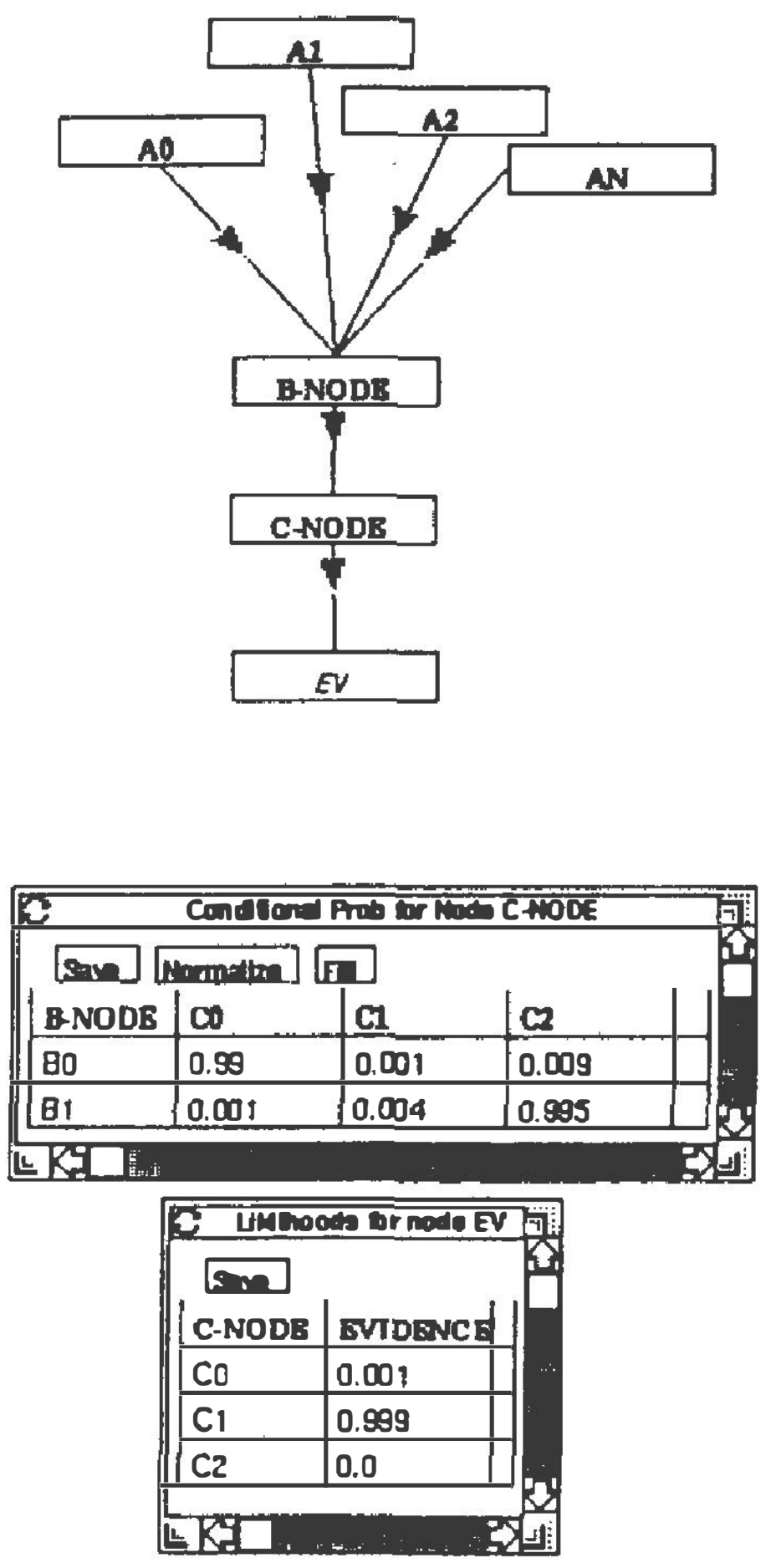

Figure 1: The Original Network

The idea is to convert the extremal likelihoods to more comparable numbers so that the random samples can be generated more uniformly among the different values.

For example, given the network and the associated conditional probability shown in Figure 1, the convergence rate of the evidence weighting technique will be very slow due to the extremal likelihood values and the conditional probabilities. Note that nodes $A0$ to $AN$ are assumed to have binary values, their prior probabilities and the conditional probability of $B$ node given them are assumed to be uniform and not given in the figure. By reversing only the arc between the evidence and its predecessor, we have integrated the evidence "partially" into the network. The resulting network and the new conditional probability is given in Figure 2. As can be seen, the evidence likelihoods now become more comparable to each other. Preliminary simulation results show that with the same evidence weighting technique, the convergence rate of the second network is at least about 100 times faster than the first one.

Note that in this example, we have only integrated the evidence "partially" into the network. This is to avoid the expensive arc reversal operation between the evidence and $B$ node. In practice, the "amount" of evidence integration would be determined by the trade-off between the fixed cost in arc reversal operations and the dynamic costs in the simulation under different convergence rates.

| | | |
|---|---|---|
| $P(A)$: | $P(a) = 0.20$ | |
| $P(B\|A)$: | $P(b\|a) = 0.80$ | $P(b\|\neg a) = 0.20$ |
| $P(C\|A)$: | $P(c\|a) = 0.20$ | $P(c\|\neg a) = 0.05$ |
| $P(D\|B,C)$: | $P(d\|b,c) = 0.80$ | $P(d\|\neg b,c) = 0.80$ |
| | $P(d\|b,\neg c) = 0.80$ | $P(d\|\neg b,\neg c) = 0.05$ |
| $P(E\|C)$: | $P(e\|c) = 0.80$ | $P(e\|\neg c) = 0.60$ |

Table 1: Probability Distributions of Example Graph

## 4 Example

In this section, the basic evidence weighting mechanisms described in Section 2, along with the "Markov blanket" algorithm, the logic sampling algorithm, and the evidential integration algorithm are simulated with the following simple example [3].

> Metastatic cancer is a possible cause of a brain tumor and is also an explanation for increased total serum calcium. In turn, either of these could possibly explain a patient falling into a coma. Severe headache is also possibly associated with a brain tumor. Assume that we are observe that a particular patient has severe headaches but not a coma.

Figure 3 shows a Bayes Network which represents these relationships. Table 1 characterizes the quantitative relationships between the variables in the network. We used the four simulation algorithms on this network to produce some anecdotal evidence for the promise of the evidence weighting mechanism.

The simulations were implemented in QUANTA, a bayesian network research and development environment which is written in CommonLisp and runs on SUN workstations. The data was obtained for five different numbers of trials per simulation run. They are: 100, 200, 500, 1000, and 2000 trials. For each setting and for each algorithm, 100 simulation runs were performed. As an assessment measure, we have used the average accumulated absolute error defined as below:

$$Error \doteq \frac{1}{N}\sum_{i=1}^{N}\sum_{x}\sum_{j} \mid p_i^e(x_j) - p(x_j) \mid \qquad (4)$$

where $x$ is a state node, $p_i^e(x_j)$ is the estimated posterior probability of the $i-th$ run for the state value $x_j$, $p(x_j)$ is the true posterior probability of $x_j$, and $N$ is the total number of simulation runs. Other assessment measures have been suggested which emphasise different characteristics of the estimate.

Figures 4 through 8 shows the results from estimating the posteriors with each of the four simulation algorithms. In Figure 4, the average accumulated absolute errors from all the nodes in the network is plotted against the number of simulation trials. It can be seen from the figure that the errors reduce as the number of trials increase. In fact, the errors decrease approximately proportional to the inverse of the square root of the number of trials (see Figure 5). Theoretically, this can be easily shown using the Central Limit Theorem. Figure 6 shows the relationship between the standard deviation of the estimates and the number of simulation trials per run, where the standard deviation is defined as the square root of the difference between the average accumulated error square and the average accumulated error. As expected, it can be seen that the deviations decrease as the number of simulation trials increase.

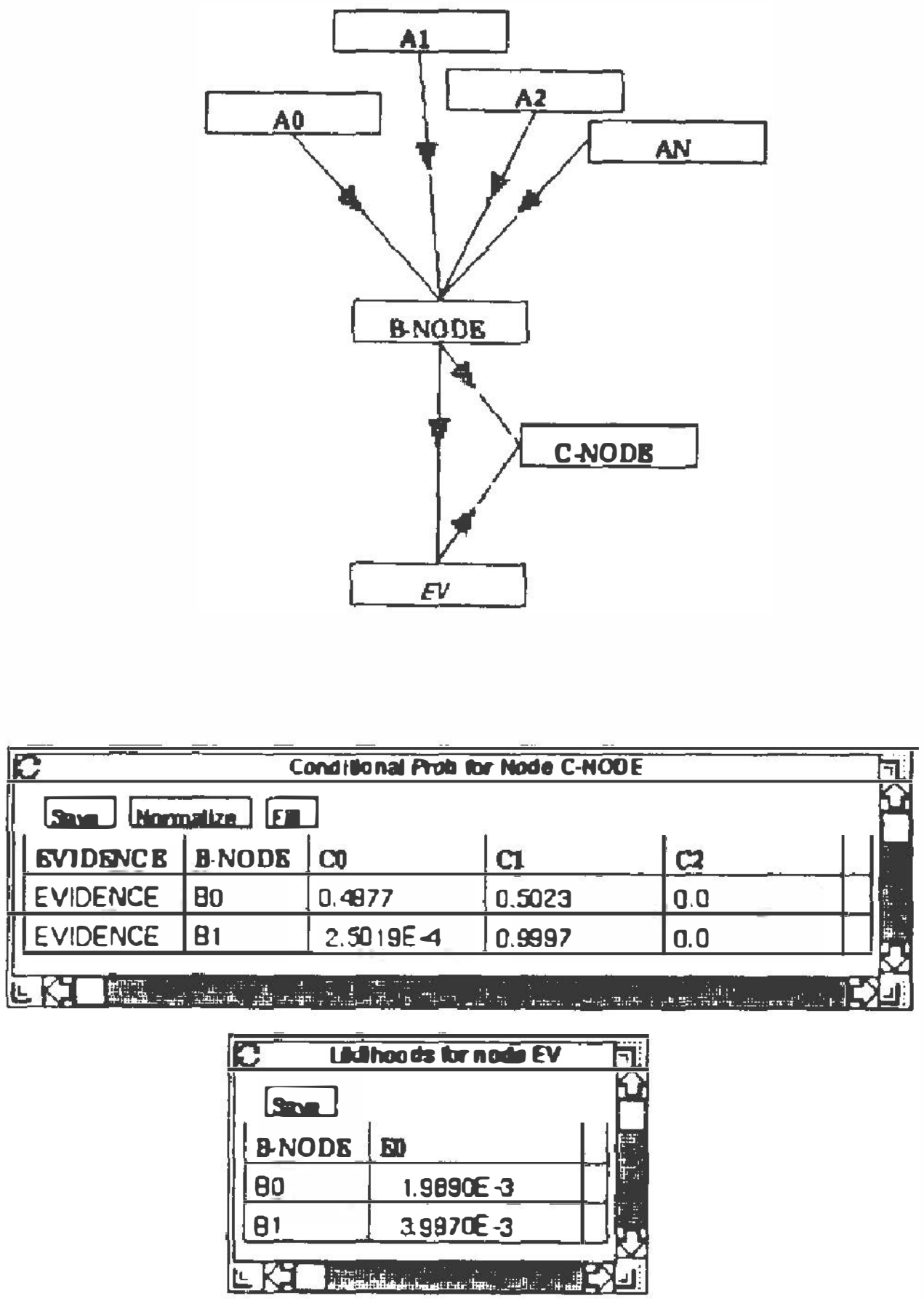

Figure 2: The Network after Integrating Evidence

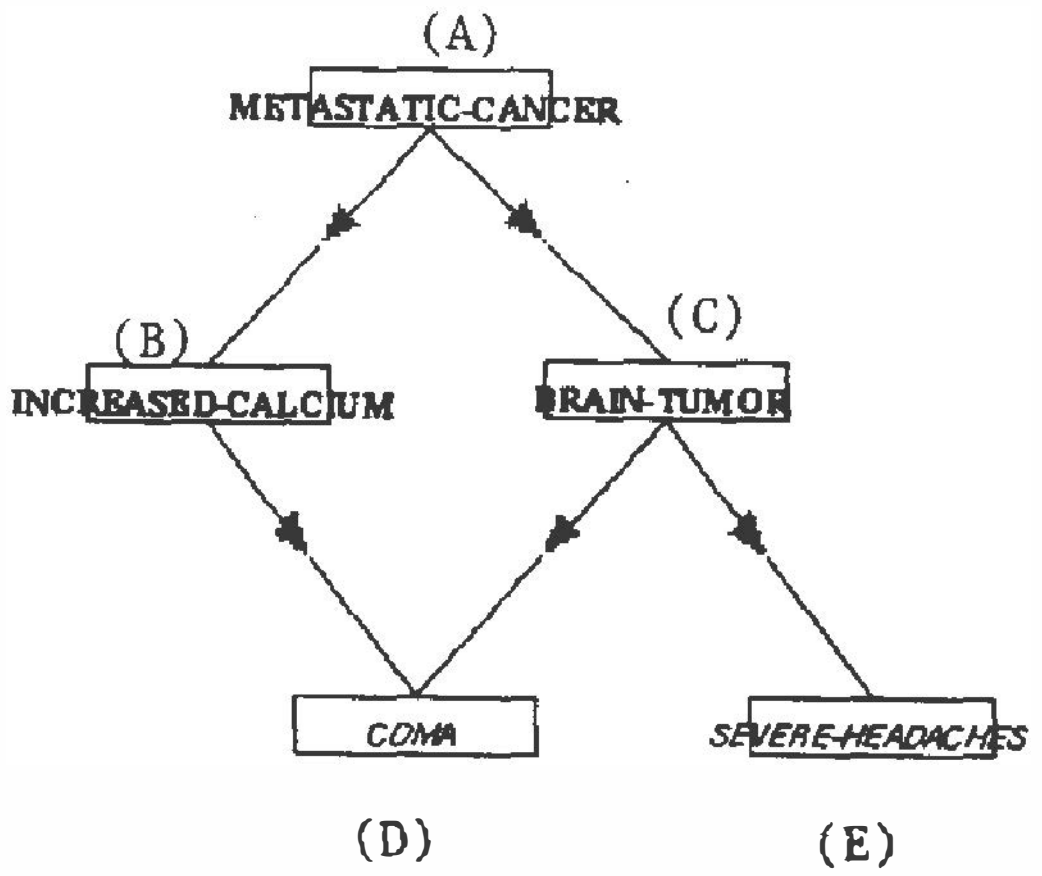

Figure 3: Example Network

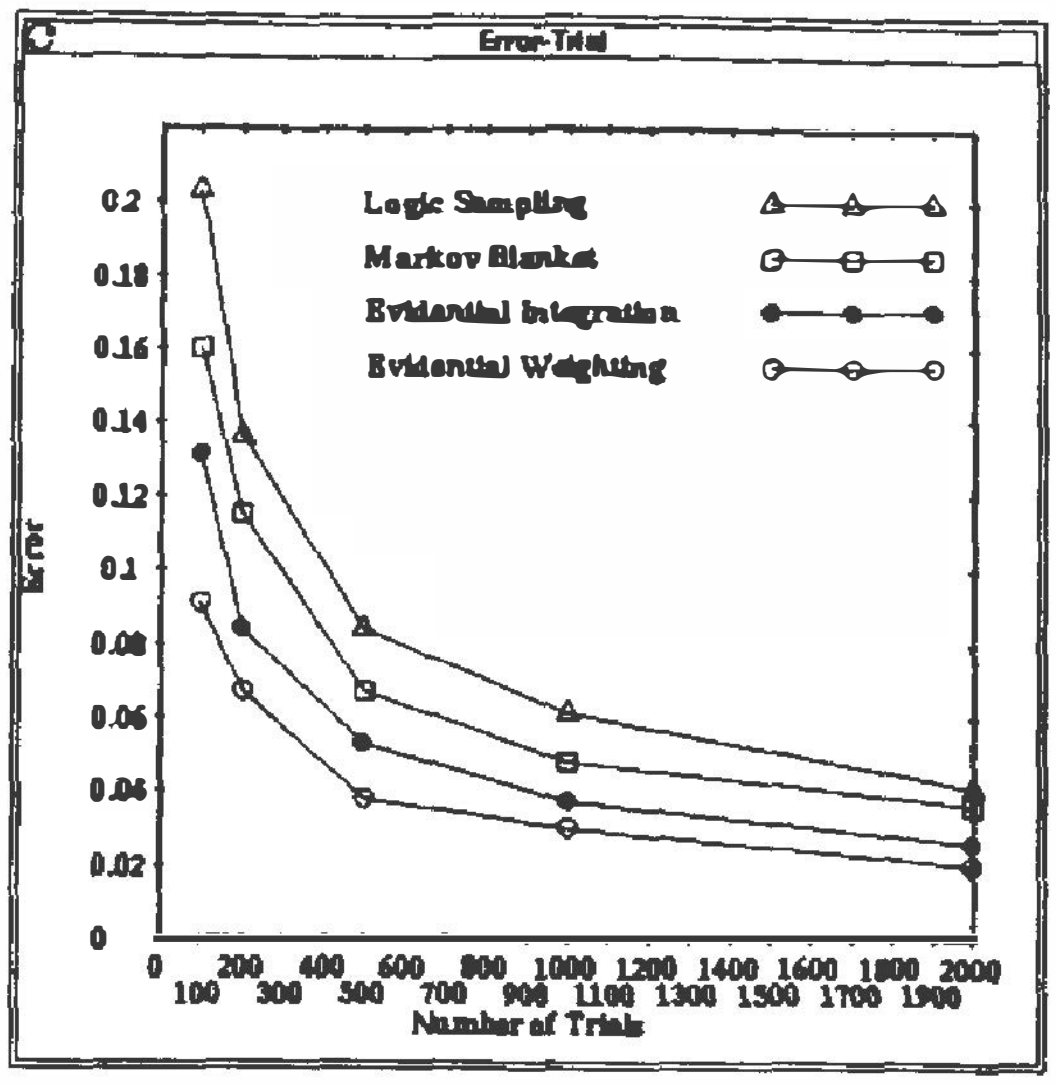

Figure 4: Error versus Number of Trials

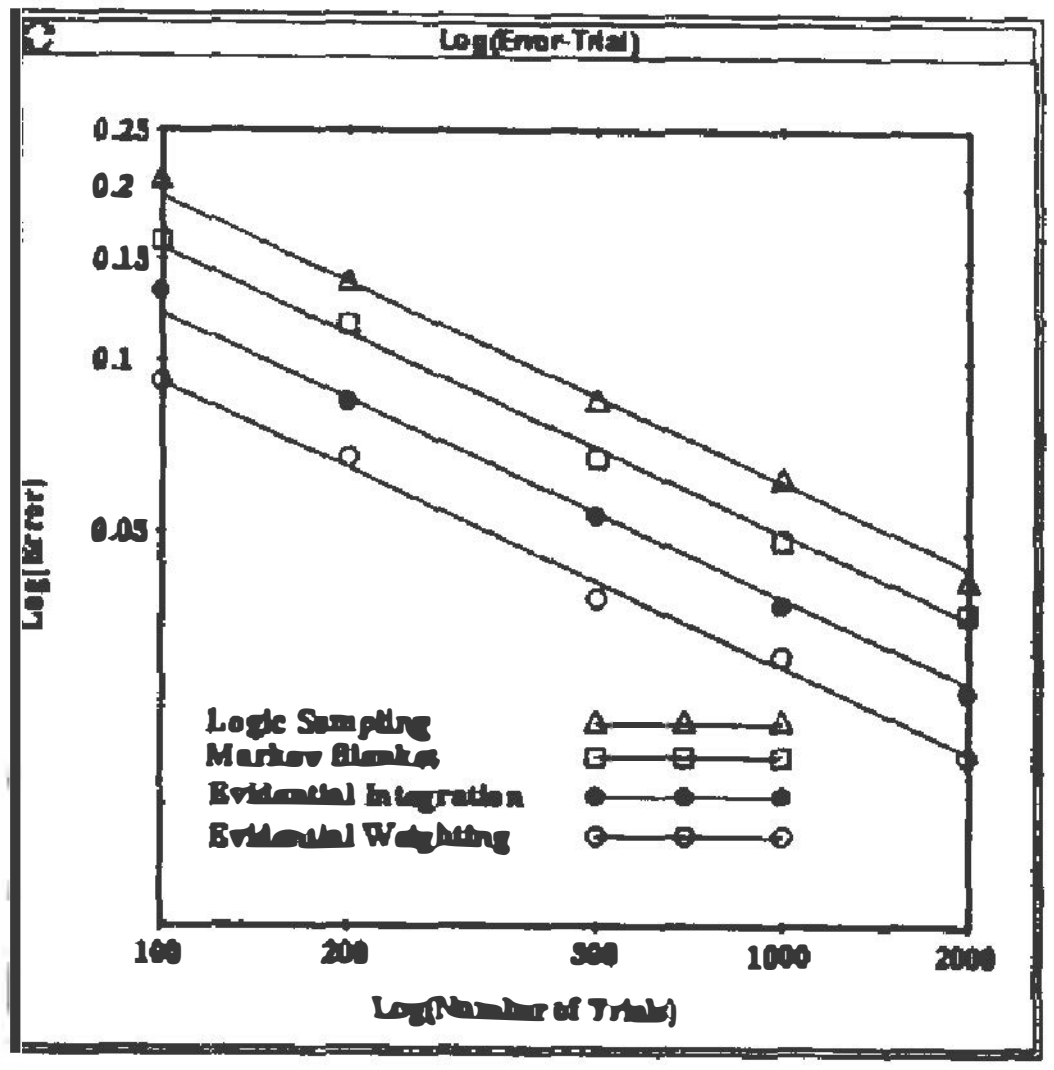

Figure 5: Log(Error) versus Log(Number of Trials)

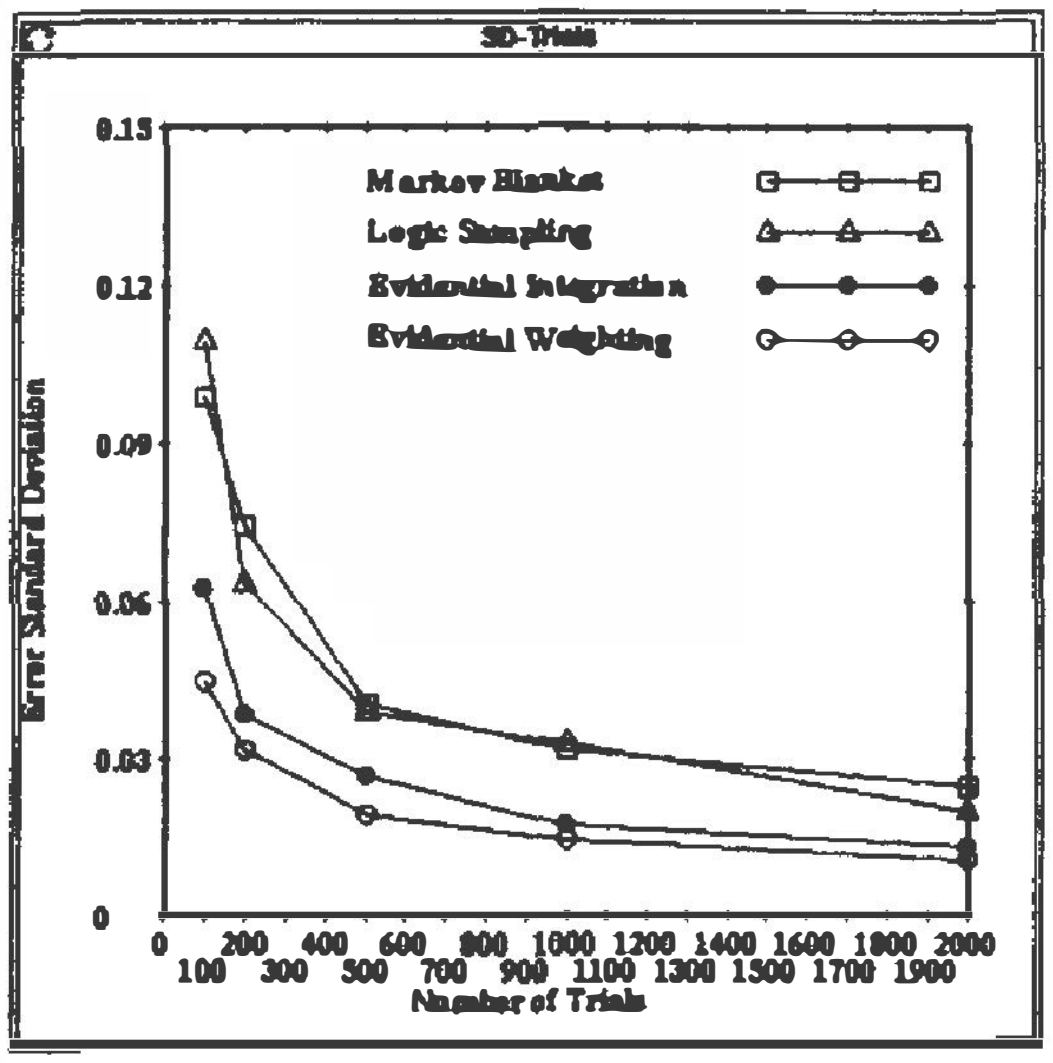

Figure 6: Standard Deviation versus Number of Trials

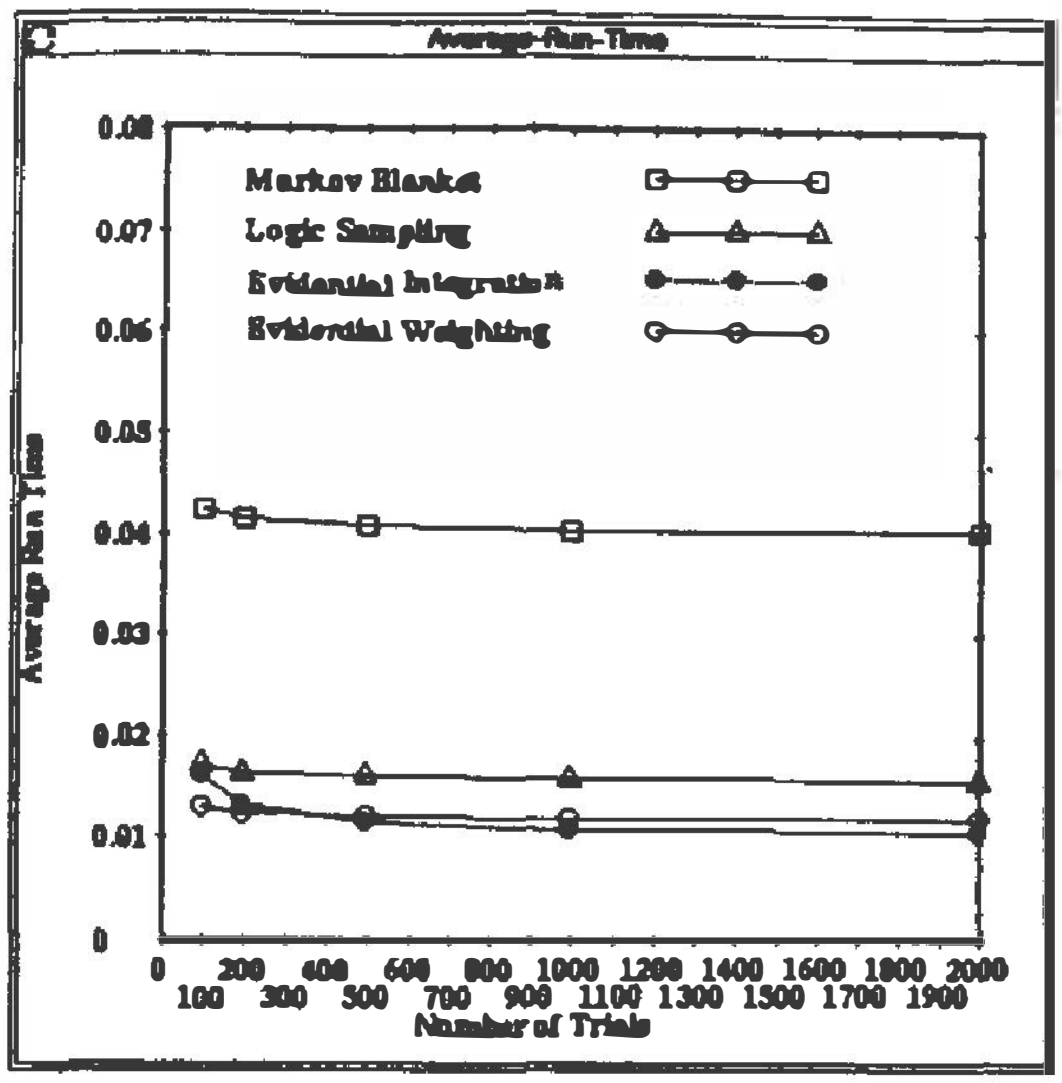

Figure 7: Average Run Time per Trial

Figure 7 shows the comparison of average run time per trial for each of the four algorithms. The effect of the fixed cost of evidential integration can be seen in the runs with smaller trials. It can also be seen that the "Markov blanket" has by far the highest cost per trial of any technique. This disparity will grow larger with the size of the network. It can be seen that the other three algorithms have very similar costs per trial. This is obvious since the algorithms have similar sampling mechanisms.

Combining Figure 4 and 7, Figure 8 shows the estimation error versus run time for the four algorithms. It can be clearly seen that for this network the evidence weighting and evidential integration mechanisms are significantly more accurate for a given amount of computation time.

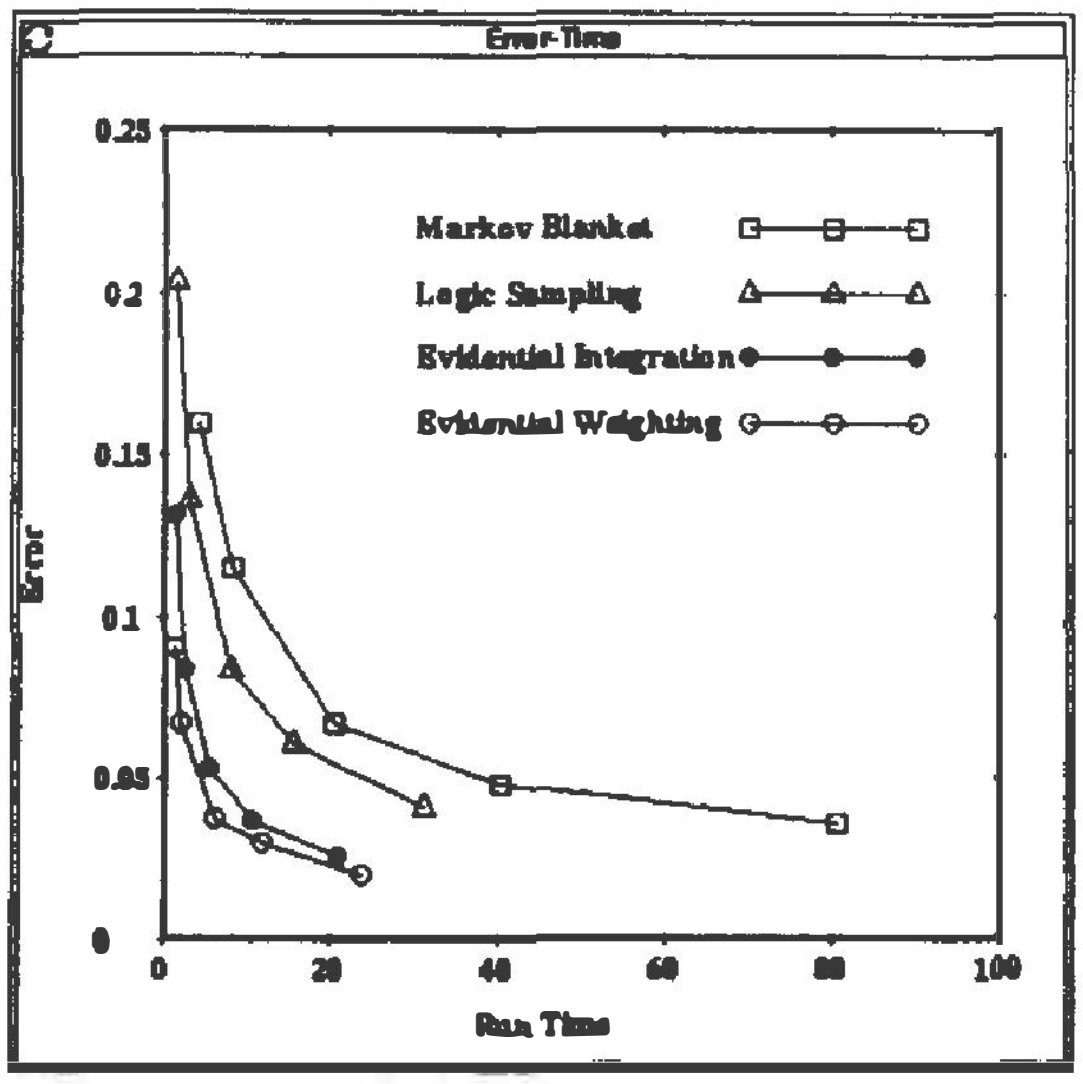

Figure 8: Error versus Run Time

## 5 Discussion

In this section, the salient features of the evidential weighting mechanism are discussed. The major advantages are three-fold. First, the mechanism have the ability to handle deterministic functions which is the major drawback of the Markov blanket approach. Since the mechanism samples only in a "causal" direction, the presence of deterministic functions will not affect their operation. Second, the evidence weighting mechanism is relatively simple. This not only allows it to be easily understood and implemented but also allows other mechanisms to be easily integrated with it. And thirdly, since all the trials are used for calculating posteriors, the mechanism seems to have good convergence properties for a large class of networks. Although the evidential integration algorithm also has these advantages, it has the fixed cost of the evidential integration step as its major disadvantage. For certain networks, the arc reversal operations used in the evidential integration process may dramatically increase the required computation and memory.

The main disadvantage of the evidence weighting algorithm is that it will converge slowly in the situations where the likelihoods for evidence are extremal. In these situations, the algorithm reduces to the logic sampling algorithm. We have shown by incorporating the evidential integration mechanism, these types of problems can be avoided. Other recent study [9] shows that combining evidence weighting with other mechanisms (e.g., Markov blanket) may also be useful. It appears from these research results that evidence weighting is a promising algorithm.

Since there seems to be a large number of performance analysis methods in use, some further research into both theoretical and simulation methods would be useful. Some new techniques for convergence analysis of simulation methods have recently been put forward [1]. Such analysis may be able to confirm the usefulness of the simulation mechanism proposed by this paper.

## 6 Conclusions

In this paper, we have presented a simple but promising mechanism for stochastic simulation, evidence weighting. The use of this mechanism appears to have many advantages. It is able to deal with deterministic variables, and the cost of each sample run is relatively low. The basic evidence weighting mechanism has the drawback of converging to the logic sampling algorithm under certain conditions. We have proposed the combination of evidential integration and evidence weighting to avoid this drawback. These results are preliminary, further research is needed in the area of convergence analysis.

Since probabilistic inference in Bayesian Networks is computationally hard, we believe no one algorithm will be able to perform optimally in every situation (e.g., time constraints, accuracy goals, network topology). Instead specialized algorithms are needed to fill "market" niches (e.g., singly-connected networks, non-exact inference), and intelligent meta-level control mechanisms are needed to match situations (e.g., network topology, network distributions and desired accuracy) with algorithms.

*Acknowledgement:* The authors wish to thank the fruitful discussion and comments from Ross Schachter, Mark Peot and Shozo Mori.